\documentclass[conference]{IEEEtran}
\usepackage{cite}
\usepackage{amsmath,amssymb,amsfonts}
\usepackage{algorithmic}
\usepackage{graphicx}
\usepackage{textcomp}
\usepackage{xcolor}
\usepackage{colortbl}
\usepackage{tabularx}
\usepackage{relsize}
\usepackage{pifont}
\usepackage{booktabs} 
\usepackage{multirow}
\usepackage{multicol}
\usepackage{adjustbox}
\usepackage{float}
\usepackage{changepage}
\usepackage{url}
\usepackage{enumitem}
\usepackage[capitalize]{cleveref}

\usepackage{graphicx}
\usepackage{amsmath}
\usepackage{amssymb}
\usepackage{booktabs}
\usepackage{algorithm}
\usepackage{algorithmic}
\usepackage{amsmath}
\usepackage{amsmath}
\usepackage{amssymb}

\usepackage{booktabs}
\usepackage{siunitx}
\usepackage{caption}
\usepackage{graphicx}  % For \resizebox
\usepackage{booktabs}  % For professional looking tables

\ifCLASSINFOpdf
\else
\fi
\begin{document}
\title{ZeroReg3D: A Zero-shot Registration Pipeline for 3D Consecutive Histopathology Image Reconstruction}

\author{Juming Xiong$^{1}$, Ruining Deng$^{2}$, Jialin Yue$^{3}$, Siqi Lu$^{3}$, Junlin Guo$^{1}$, Marilyn Lionts$^{3}$, Tianyuan Yao$^{3}$, Can Cui$^{3}$,Junchao Zhu$^{3}$, Chongyu Qu$^{1}$,  Mengmeng Yin$^{4}$, Haichun Yang$^{4}$, Yuankai Huo$^{1,3}$*}

\author{\IEEEauthorblockN{Juming Xiong}
\IEEEauthorblockA{Department of Electrical and \\ Computer Engineering\\
Vanderbilt University, TN, USA}
\and
\IEEEauthorblockN{Ruining Deng}
\IEEEauthorblockA{Department of Radiology\\ Weill Cornell Medicine, NY, USA}
\and
\IEEEauthorblockN{Jialin Yue}
\IEEEauthorblockA{Department of Electrical and \\ Computer Engineering\\
Vanderbilt University, TN, USA}
\and
\IEEEauthorblockN{Siqi Lu}
\IEEEauthorblockA{Department of Electrical and \\ Computer Engineering\\
Vanderbilt University, TN, USA}
\and
\IEEEauthorblockN{Junlin Guo}
\IEEEauthorblockA{Department of Electrical and Computer Engineering\\
Vanderbilt University, TN, USA}
\and
\IEEEauthorblockN{Marilyn Lionts}
\IEEEauthorblockA{Department of Computer Science\\
Vanderbilt University, TN, USA}
\and
\IEEEauthorblockN{Tianyuan Yao}
\IEEEauthorblockA{Department of Computer Science\\
Vanderbilt University, TN, USA}
\and
\IEEEauthorblockN{Can Cui}
\IEEEauthorblockA{Department of Computer Science\\
Vanderbilt University, TN, USA}
\and
\IEEEauthorblockN{Junchao Zhu}
\IEEEauthorblockA{Department of Computer Science\\
Vanderbilt University, TN, USA}
\and
\IEEEauthorblockN{Chongyu Qu}
\IEEEauthorblockA{Department of Electrical and \\ Computer Engineering\\
Vanderbilt University, TN, USA}
\and
\IEEEauthorblockN{Yuechen Yang}
\IEEEauthorblockA{Department of Computer Science\\
Vanderbilt University, TN, USA}
\and
\IEEEauthorblockN{MengMeng Yin}
\IEEEauthorblockA{Department of Pathology, Microbiology,\\ and Immunology\\ Vanderbilt University Medical Center, TN, USA}
\and
\IEEEauthorblockN{Haichun Yang}
\IEEEauthorblockA{Department of Pathology, Microbiology,\\ and Immunology\\ Vanderbilt University Medical Center, TN, USA}
\and
\IEEEauthorblockN{Yuankai Huo}
\IEEEauthorblockA{Department of Computer Science\\Department of Electrical and Computer Engineering\\
Vanderbilt University, TN, USA\\
$\star$ Corresponding Author}}


%Telephone: (800) 555--1212, Fax: (888) 555--1212}


\maketitle

\begin{abstract}
Histological analysis plays a crucial role in understanding tissue structure and pathology. While recent advancements in registration methods have improved 2D histological analysis, they often struggle to preserve critical 3D spatial relationships, limiting their utility in both clinical and research applications. Specifically, constructing accurate 3D models from 2D slices remains challenging due to tissue deformation, sectioning artifacts, variability in imaging techniques, and inconsistent illumination. Deep learning-based registration methods have demonstrated improved performance but suffer from limited generalizability and require large-scale training data. In contrast, non-deep-learning approaches offer better generalizability but often compromise on accuracy. In this study, we introduced ZeroReg3D, a novel zero-shot registration pipeline tailored for accurate 3D reconstruction from serial histological sections. By combining zero-shot deep learning-based keypoint matching with optimization-based affine and non-rigid registration techniques, ZeroReg3D effectively addresses critical challenges such as tissue deformation, sectioning artifacts, staining variability, and inconsistent illumination without requiring retraining or fine-tuning. The code has been made publicly available at \color{blue}{\url{https://github.com/hrlblab/ZeroReg3D}}
\end{abstract}

% \keyword{medical images, anomaly detection, digital pathology} 

% \fi
\IEEEpeerreviewmaketitle

\section{Introduction}
\begin{figure*}[t]
\centering 
\includegraphics[width=1\linewidth]{{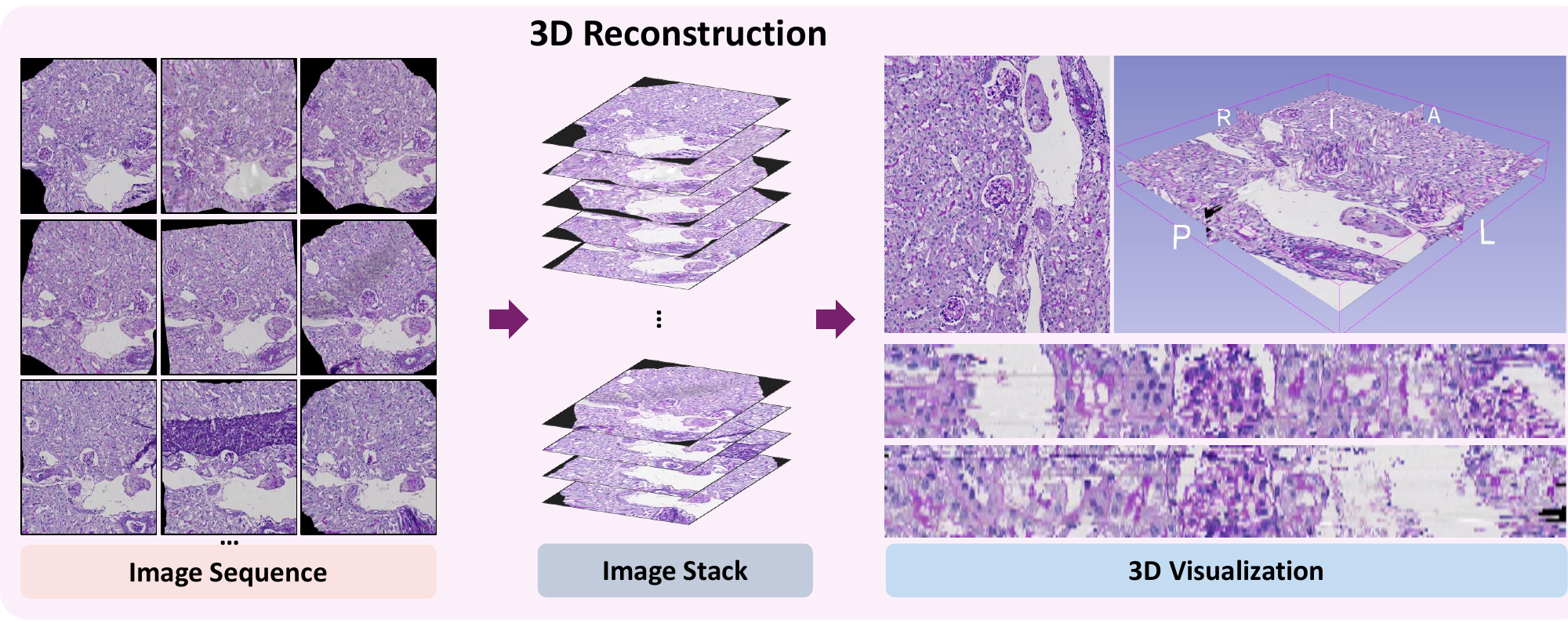}}
\caption{\textbf{Overview.} This figure shows a reconstructed 3D volume after aglignment. The image sequence was stacked and subjected to 3D visualization to provide a comprehensive view.} 
\label{overview} 
\end{figure*}

Histology plays an essential role in both clinical diagnosis and biomedical research\cite{xiong2024deep,Zhu_2025_CVPR}. Scientists and clinicians alike analyze tissue slides to identify key cellular structures and pathological features\cite{deng2024hats,deng2024prpseg}. In recent years, an increasing number of deep learning-based methods have been proposed to aid in tissue analysis and disease diagnosis\cite{Litjens, Kather, Brinker, Kiehl, guo2024assessment,guo2024good,zhu2025magnet,xiong2024circle,yue2025weighted}. However, these methods, which rely on two-dimensional (2D) slices, fail to capture the intricate, three-dimensional (3D) spatial architecture of tissues present in histological samples, leading to a loss of information surrounding disease mechanisms \cite{Chen, Merz}and potential therapeutic targets \cite{Geng}. By visualizing the intricate spatial architecture of tissues in three dimensions, researchers and clinicians can gain deeper insights into disease mechanisms, progression, and potential therapeutic targets. However, the transition from 2D histological slices to comprehensive 3D models remains a significant challenge across multiple domains of pathology as shown in~\Cref{overview}.

Although significant progress has been made toward accurate 3D reconstruction of 2D consecutive histopathology images \cite{Pichat}, numerous obstacles persist in the field, including tissue deformation, sectioning artifacts, variability in staining techniques, and inconsistent illumination across sections \cite{saalfeld2012elastic}. Tissue deformation arises from biological variability and the mechanical stresses involved in slicing, leading to misalignment between consecutive sections.\cite{auto3dreg2015} Sectioning artifacts, such as tears or folds, further complicate the registration process, while heterogeneous staining and differing illumination conditions can obscure anatomical features necessary for accurate registration \cite{lotz2021comparison}. 

In this paper, we introduce ZeroReg3D, a hybrid registration pipeline that integrates zero-shot deep learning and non-deep-learning registration methods for 3D reconstruction using routinely collected 2D WSIs. The proposed pipeline does not require retraining or finetuning by employing zero-shot XFeat \cite{potje2024xfeatacceleratedfeatureslightweight} key point extraction and matching, as well as affine registration and B-spline based deformable registration. Our 3D reconstruction method has been evaluated by mice whole kidney sections and human needle biopsy sections, demonstrating that our method outperforms existing registration strategies in both accuracy and robustness. The contributions of this paper are threefold:

$\bullet$ \textbf{Holistic zero-shot deep learning and non-deep-learning design}: Our method does not require additional retraining or finetuning on new data. The zero-shot deep learning based keypoint matching ensures broad applicability and facilitates easy integration across diverse histological datasets.

$\bullet$ \textbf{Comprehensive studies}: Our 3D reconstruction method has been evaluated by mice whole kidney sections and human needle biopsy sections, covering the two prevalent biopsy formats and different spices.

$\bullet$ \textbf{Open-source deployment}: The entire pipeline has been released as an open-source package, which is publicly available at \url{https://github.com/hrlblab/ZeroReg3D}.

\begin{figure}
\centering 
\includegraphics[width=1\linewidth]{{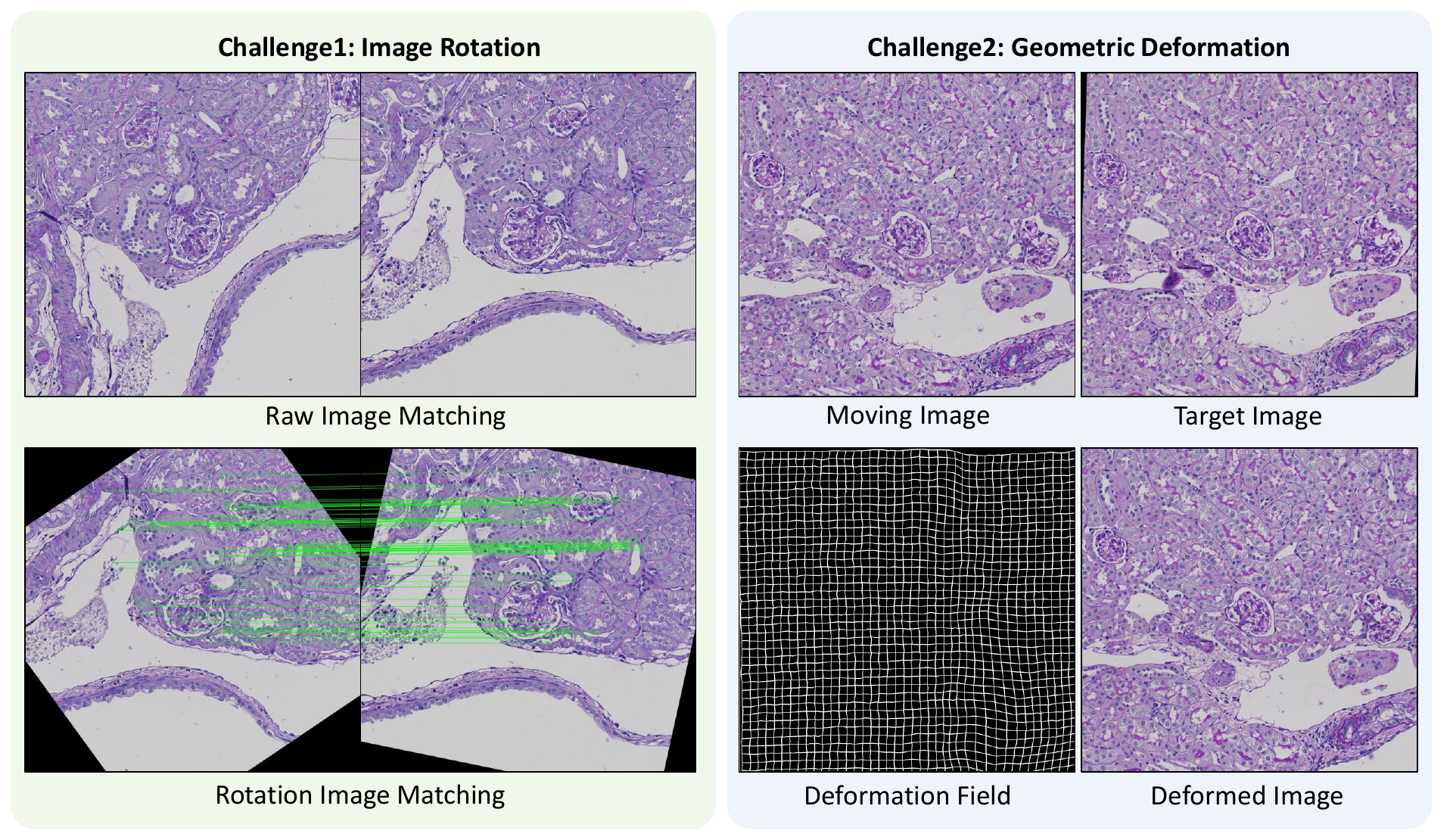}}
\caption{\textbf{Challenges.} This figure showed two major challenges in the registration process. The left panel showed the image rotation problem will affect the keypoint matching result. The right panel showed the geometric deformation between moving image and target image that need use non-rigid registration to solve help solving this problem.} 
\label{challenge} 
\end{figure}

\begin{figure*}[t]
\centering 
\includegraphics[width=1\linewidth]{{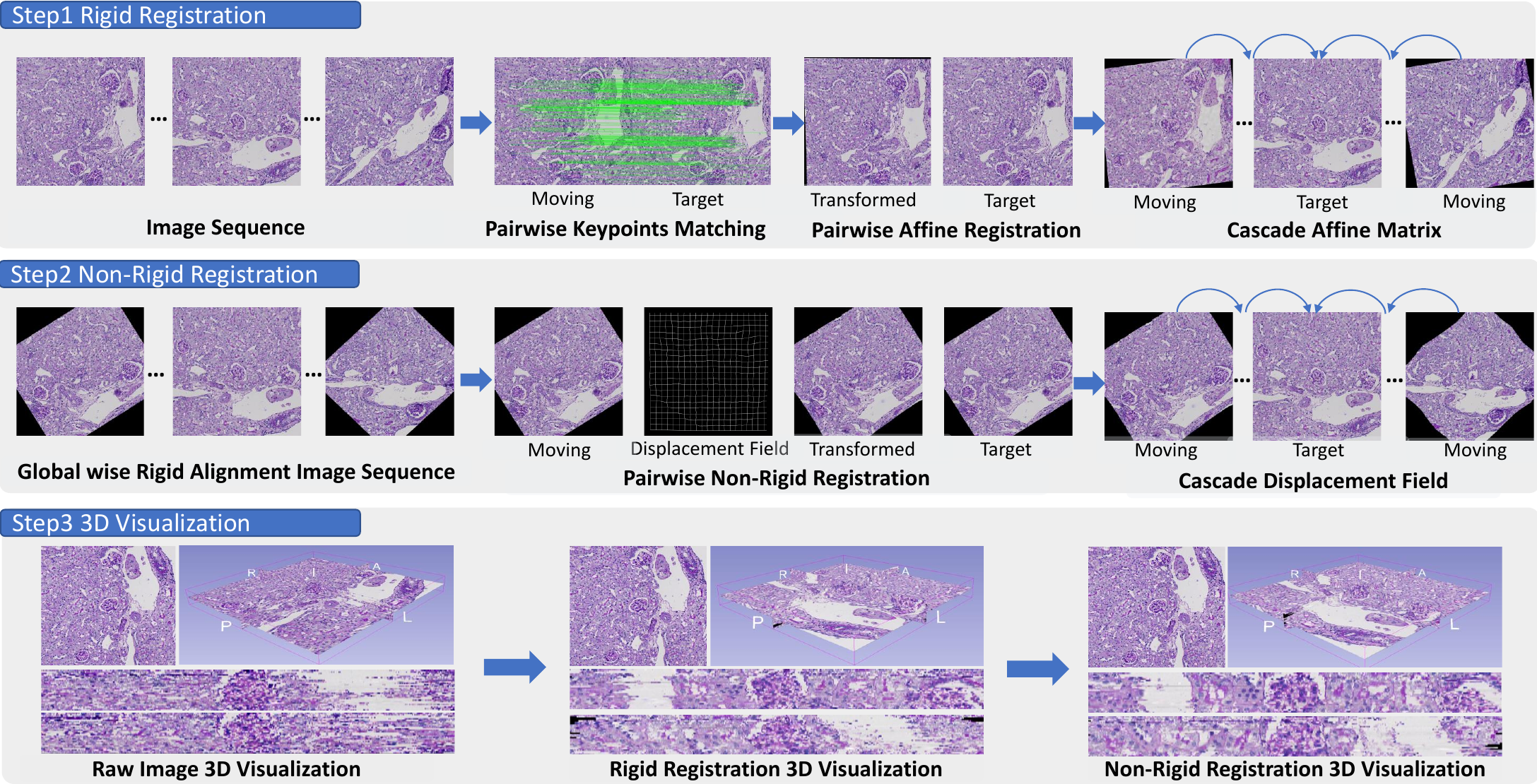}}
\caption{\textbf{Method.} This figure shows the entire pipeline and the visualization difference between raw image sequence, rigid registration image sequence and non-rigid registration image sequence.} 
\label{method} 
\end{figure*}

%%%%%%%%%%%%%%%%%%%%%%%%%%%%%%%%%%%%%%%%%%
\section{Related work}

\subsection{Deep Learning-Based Registration}
Deep-learning based methods leverage neural networks to learn robust feature representations that facilitate image registration by directly predicting correspondences between images. For instance, methods such as SuperGlue\cite{DBLP:journals/corr/abs-1911-11763} and OmniGlue\cite{jiang2024omnigluegeneralizablefeaturematching} utilize graph neural network architectures to refine feature matching by considering the contextual relationships among detected keypoints, resulting in more reliable global registration even under challenging conditions. Transformer-based approaches like LoFTR\cite{DBLP:journals/corr/abs-2104-00680} establish dense, context-aware correspondences across image pairs without relying on explicit keypoint detection, while CNN-based frameworks like XFeat\cite{potje2024xfeatacceleratedfeatureslightweight} offer efficient feature extraction, particularly suited for resource-limited scenarios. Importantly, although these deep learning methods vary in architecture, they typically focus on estimating a single global transformation (e.g., a homography) for image registration, as demonstrated by DeTone et al.\cite{DeTone2016DeepIH}.

\subsection{Non-Deep Learning Based Registration}
Non-deep learning-based methods form the traditional foundation of image registration by relying on handcrafted features and well-established mathematical models. These techniques include intensity-based approaches that utilize similarity metrics such as mutual information, cross-correlation, and structural similarity \cite{466930, 1660446, 1284395} to determine the optimal registration between images. In addition, feature-based methods employing algorithms like SIFT and ORB \cite{Lowe2004DistinctiveIF, 6126544} detect and match distinctive keypoints to compute transformation parameters for global registration. When it comes to capturing local, non-rigid deformations, methods such as B-splines \cite{Rueckert1999NonrigidRU}, the Demons algorithm \cite{thirion1998image}, and Thin Plate Splines (TPS) \cite{24792} are employed to model spatially varying transformations. These approaches are often integrated into comprehensive software suites like Advanced Normalization Tools (ANTs)\cite{avants2009advanced}. 

\subsection{Registration Methods in Pathology} In pathology, specialized registration methods have emerged to address unique challenges associated with serial histological section registration, including tissue deformation, missing sections, and staining variability. Map3D \cite{deng2021map3dregistrationbasedmultiobject} employs a registration-based framework integrating deep learning for the automated identification and association of large-scale 3D glomerular structures across serial renal histology sections, employing a quality-aware strategy to overcome these challenges. Similarly, DeeperHistReg \cite{wodzinski2024deeperhistregrobustslideimages} combines deep learning feature extraction with traditional registration techniques to robustly align consecutive histopathology slides stained differently, facilitating precise registration critical for downstream pathological analyses.

%%%%%%%%%%%%%%%%%%%%%%%%%%%%%%%%%%%%%%%%%%
\section{Method}
The entire framework of the proposed 3D registration is presented in ~\Cref{method}. This pipeline consists of three sections:
(1) Keypoints extraction and matching (2) Affine registration (3) B-spline based non-rigid registration.

\subsection{Keypoints extraction and matching}

Before performing the affine registration, we apply rotation preprocessing to the moving image to maximize the number of valid keypoint matching pairs with the fixed image. This preprocessing step enhances the registration accuracy by ensuring that the moving image is oriented optimally relative to the fixed image, thereby facilitating a higher number of accurate keypoint matches as shown in ~\Cref{challenge}.

Rigid registration serves as the initial step in our registration pipeline, addressing global misalignments between consecutive histological sections. This stage ensures that the overall orientation, scale, and position of the tissue slices are consistent before finer adjustments are made. We employed XFeat\cite{potje2024xfeatacceleratedfeatureslightweight} as our image matching method for initial registration technique for the initial registration due to its speed and effectiveness.

\subsection{Affine Registration}

Building upon the robust keypoint matches obtained through our rotation invariant keypoint matching framework, we employ affine registration to achieve precise registration of consecutive histological sections. Affine registration models the transformation between two image slices as a combination of linear transformations (rotation, scaling, shearing) and translation, allowing for both global registration and local adjustments to accommodate minor deformations.

\begin{equation}
\begin{bmatrix}
x' \\
y' \\
1
\end{bmatrix}
=
\begin{bmatrix}
a_{11} & a_{12} & t_x \\
a_{21} & a_{22} & t_y \\
0 & 0 & 1
\end{bmatrix}
\begin{bmatrix}
x \\
y \\
1
\end{bmatrix}
\label{eq:affine_transformation}
\end{equation}

where:
\((x, y)\) are the coordinates in the moving image \(I_{t+1}\),
\((x', y')\) are the coordinates in the fixed image \(I_t\),
\(a_{11}, a_{12}, a_{21}, a_{22}\) are elements of the affine transformation matrix \(\mathbf{A}\), \(t_x, t_y\) are components of the translation vector \(\mathbf{t}\).

To estimate the affine transformation parameters, we follow these steps:

1. Keypoint Detection and Description: Keypoints are detected in both the fixed image \(I_t\) and the moving image \(I_{t+1}\) by using XFeat\cite{potje2024xfeatacceleratedfeatureslightweight} after rotation preprocessing. Descriptors are computed for each keypoint to capture the local image structure.

2. Keypoint Matching: The descriptors from the fixed and moving images are matched to establish correspondences between keypoints. This results in a set of matched pairs \(\mathcal{M} = \{ (p_i, q_i) \}\), where \(p_i\) is a keypoint in the fixed image and \(q_i\) is the corresponding keypoint in the moving image.

3. Outlier Rejection using RANSAC: To eliminate erroneous matches, we apply the RANdom SAmple Consensus (RANSAC) algorithm. RANSAC iteratively selects random subsets of matches, estimates the affine transformation, and computes the number of inliers that agree with this model. The transformation with the highest number of inliers is selected.

4. Affine Transformation Estimation: Using the set of inlier correspondences \(\mathcal{I} \subset \mathcal{M}\), we estimate the affine transformation parameters by solving the following system of equations for each inlier pair \((p_i, q_i)\):

\begin{algorithm}
\caption{Consecutive Affine Registration using Keypoint Matching and RANSAC}
\begin{algorithmic}[1]
\REQUIRE Image sequence \(I_1, I_2, \dots, I_N\)
\FOR{\(t = 1\) to \(N-1\)}
    \STATE \textbf{Set} Fixed image \(\leftarrow I_t\)
    \STATE \textbf{Set} Moving image \(\leftarrow I_{t+1}\)
    \STATE \textbf{Detect} keypoints and compute descriptors in Fixed and Moving images
    \STATE \textbf{Match} keypoints between Fixed and Moving images to obtain set \(\mathcal{M}\)
    \STATE \textbf{Apply} RANSAC to \(\mathcal{M}\) to filter out outliers, resulting in inlier set \(\mathcal{I}\)
    \STATE \textbf{Estimate} affine transformation parameters \((\mathbf{A}, \mathbf{t})\) using \(\mathcal{I}\)
    \STATE \textbf{Transform} Moving image using affine transformation: \(\mathbf{x}' = \mathbf{A}\mathbf{x} + \mathbf{t}\)
    \STATE \textbf{Optional:} Store transformation parameters or registered image
\ENDFOR
\end{algorithmic}
\end{algorithm}

\subsection{B-spline Based Non-Rigid Registration}
While rigid registration effectively addresses global misalignments, biological tissues often undergo local deformations that require more flexible transformation models. To capture and correct these local distortions, we employ B-spline based non-rigid registration, which allows for spatially varying transformations, enabling precise registration of complex anatomical structures by adapting to subtle morphological changes between sections as shown in ~\Cref{challenge}.

The B-spline based non-rigid registration models the deformation field using a grid of control points overlaid on the image domain. The displacement of each pixel is calculated based on the displacements of the surrounding control points using B-spline basis functions. The transformation \( \mathbf{T}(\mathbf{x}) \) of a point \( \mathbf{x} = (x, y) \) is given by:

\begin{equation}
\mathbf{T}(\mathbf{x}) = \mathbf{x} + \sum_{i=0}^{n} \sum_{j=0}^{m} \mathbf{c}_{i,j} \, B_i(u) \, B_j(v)
\label{eq:b_spline_transformation}
\end{equation}

Where, \( \mathbf{c}_{i,j} \) are the displacement vectors (control point coefficients) at control point \( (i, j) \). The functions \( B_i(u) \) and \( B_j(v) \) are the B-spline basis functions of degree \( k \). The variables \( u \) and \( v \) are the normalized coordinates with respect to the control point grid. The parameters \( n \) and \( m \) denote the number of control points in the \( x \) and \( y \) directions, respectively.

The B-spline basis functions are defined recursively, providing smooth and continuous transformations suitable for modeling biological tissue deformations.

\subsubsection{Loss Function}

To optimize the transformation parameters \( \{\mathbf{c}_{i,j}\} \), we used a combined loss function that incorporates both image similarity and transformation regularization:

\begin{equation}
\mathcal{L} = \mathcal{L}_{\text{NCC}} + \lambda \mathcal{L}_{\text{reg}}
\label{eq:combined_loss}
\end{equation}

where \( \mathcal{L}_{\text{NCC}} \) is the local Normalized Cross-Correlation (NCC) loss measuring the similarity between the fixed image \( I_t \) and the transformed moving image \( I_{t+1} \). The term \( \mathcal{L}_{\text{reg}} \) is the regularization loss enforcing smoothness in the displacement field. The parameter \( \lambda \) is a weighting factor that balances the contribution of the regularization term.

% \paragraph{Local Normalized Cross-Correlation Loss}

The local NCC loss is defined as:

\begin{equation}
\mathcal{L}_{\text{NCC}} = - \sum_{\mathbf{x} \in \Omega} 
\frac{ S_{\text{num}}(\mathbf{x}) }{ \sqrt{ S_{\text{den1}}(\mathbf{x}) } \sqrt{ S_{\text{den2}}(\mathbf{x}) } }
\label{eq:ncc_loss_simple}
\end{equation}

where:

\begin{equation}
S_{\text{num}}(\mathbf{x}) = \sum_{\mathbf{r} \in \mathcal{N}(\mathbf{x})} 
\left[ I_t(\mathbf{r}) - \mu_t(\mathbf{x}) \right]
\left[ I_{t+1}\left( \mathbf{T}(\mathbf{r}) \right) - \mu_{t+1}(\mathbf{x}) \right]
\label{eq:s_num}
\end{equation}

\begin{equation}
S_{\text{den1}}(\mathbf{x}) = \sum_{\mathbf{r} \in \mathcal{N}(\mathbf{x})} 
\left[ I_t(\mathbf{r}) - \mu_t(\mathbf{x}) \right]^2
\label{eq:s_den1}
\end{equation}

\begin{equation}
S_{\text{den2}}(\mathbf{x}) = \sum_{\mathbf{r} \in \mathcal{N}(\mathbf{x})} 
\left[ I_{t+1}\left( \mathbf{T}(\mathbf{r}) \right) - \mu_{t+1}(\mathbf{x}) \right]^2
\label{eq:s_den2}
\end{equation}

In these equations, \( \Omega \) is the image domain, and \( \mathcal{N}(\mathbf{x}) \) denotes a local neighborhood around point \( \mathbf{x} \). The terms \( \mu_t(\mathbf{x}) \) and \( \mu_{t+1}(\mathbf{x}) \) are the local mean intensities of the fixed and moving images within \( \mathcal{N}(\mathbf{x}) \), respectively.

%%%%%%%%%%%%%%%%%%%%%%%%%%%%%%%%%%%%%%%%%%%%%%%%%%%%%%%%%%%%%%%%%%%%%
The regularization loss \( \mathcal{L}_{\text{reg}} \) is designed to encourage smoothness in the 2D displacement field \( \mathbf{u}(\mathbf{x}) \) by penalizing the squared differences of the displacement vectors between neighboring pixels, scaled relative to the image size. This diffusion regularization term is similar to VoxelMorph\cite{DBLP:journals/corr/abs-1809-05231} defined as:

\begin{equation}
\mathcal{L}_{\text{reg}} = \frac{1}{2} \left( \mathbb{E}_{\mathbf{x}} \left[ s_x^2 \left\| \delta_x \mathbf{u}(\mathbf{x}) \right\|^2 \right] + \mathbb{E}_{\mathbf{x}} \left[ s_y^2 \left\| \delta_y \mathbf{u}(\mathbf{x}) \right\|^2 \right] \right)
\label{eq:reg_loss_2d}
\end{equation}

where:

\begin{equation}
\delta_x \mathbf{u}(\mathbf{x}) = \mathbf{u}(x+1, y) - \mathbf{u}(x, y)
\label{eq:delta_x}
\end{equation}

\begin{equation}
\delta_y \mathbf{u}(\mathbf{x}) = \mathbf{u}(x, y+1) - \mathbf{u}(x, y)
\label{eq:delta_y}
\end{equation}

In these equations: \( \mathbf{u}(\mathbf{x}) = [u_x(x, y), u_y(x, y)] \) is the displacement vector at position \( \mathbf{x} = (x, y) \). \( \delta_x \mathbf{u}(\mathbf{x}) \) and \( \delta_y \mathbf{u}(\mathbf{x}) \) are the finite differences of the displacement field along the \( x \) and \( y \) directions, respectively. \( s_x = N_x \) and \( s_y = N_y \) are scaling factors equal to the number of pixels in the \( x \) and \( y \) dimensions (i.e., the image size in each dimension). \( \mathbb{E}_{\mathbf{x}} [ \cdot ] \) denotes the expectation (mean) over all valid positions \( \mathbf{x} \) in the image domain \( \Omega \). \( \left\| \cdot \right\|^2 \) denotes the squared Euclidean norm.

% -------------------------------------------

\begin{algorithm}
\caption{B-spline Based Non-Rigid Registration}
\begin{algorithmic}[1]
\REQUIRE Fixed image \( I_t \), Moving image \( I_{t+1} \), control point grid \( \{\mathbf{c}_{i,j}\} \), weighting factor \( \lambda \), learning rate \( \alpha \), maximum iterations \( N_{\text{max}} \)
\STATE Initialize control point displacements \( \mathbf{c}_{i,j} \leftarrow \mathbf{0} \) for all \( i, j \)
\FOR{iteration = 1 to \( N_{\text{max}} \)}
    \STATE Compute the transformation \( \mathbf{T}(\mathbf{x}) \) using Equation (\ref{eq:b_spline_transformation})
    \STATE Warp the moving image: \( I_{t+1}(\mathbf{T}(\mathbf{x})) \)
    \STATE Compute the loss \( \mathcal{L} \) using Equations (\ref{eq:ncc_loss_simple}) and (\ref{eq:combined_loss})
    \STATE Compute gradients \( \frac{\partial \mathcal{L}}{\partial \mathbf{c}_{i,j}} \) analytically or via automatic differentiation
    \STATE Update control points: \( \mathbf{c}_{i,j} \leftarrow \mathbf{c}_{i,j} - \alpha \frac{\partial \mathcal{L}}{\partial \mathbf{c}_{i,j}} \)
    \IF{ \( |\mathcal{L}^{\text{new}} - \mathcal{L}^{\text{old}}| < \epsilon \) }
        \STATE \textbf{Break}
    \ENDIF
\ENDFOR
\STATE \textbf{Output:} Optimized transformation \( \mathbf{T}(\mathbf{x}) \)
\end{algorithmic}
\end{algorithm}

%%%%%%%%%%%%%%%%%%%%%%%%%%%%%%%%%%%%%%%%%
\section{Experiments}

\subsection{Data}

\subsubsection{Mice Datasets}
In our study, we employed kidney tissue sections from both healthy and db/db diabetic mice to evaluate our computational models. Specifically, we analyzed 29 consecutive 2 $\mu m$-thick sections from normal mice and 39 consecutive 2 $\mu m$-thick sections from db/db diabetic mice. Each section was imaged at a high resolution of 40$\times$ magnification to capture detailed histological features. The resulting images were subsequently divided into patches of size 2048 $\times$ 2048 pixels to facilitate scalable and efficient processing within our experiments.

\subsubsection{Human Datasets}
We have 20 human kidney biopsy cases, each containing 13 consecutive whole slide needle biopsy images. Each slide includes 2 to 5 tissue sections, with each section being 2 $\mu m$ thick and captured at 40$\times$ magnification. Each case undergoes sequential staining with three different stains—H\&E first, followed by PAS and then Jones—repeated three times, with two unstained slides placed between each staining cycle. This results in the following sequence: H\&E, PAS, Jones, two unstained slides, H\&E, PAS, Jones, two unstained slides, and H\&E, PAS, Jones. To facilitate processing while maintaining detail, we downsample the whole slide images by a factor of two, making them more manageable without significant loss of information.

\subsection{Evaluation Metrics}

To quantitatively assess the performance of our registration method, we employ several evaluation metrics that measure the registration accuracy between the registered images and the ground truth landmarks.

%%%%%%%%%%%%%%%%%%%%%%%%%%%%%%%%%%%%%%%%%%%%%%%%%%%%%%%

\noindent\textbf{Relative Target Registration Error (rTRE)} We followed evaluation metric in Borovec's work\cite{9058666}. The Relative Target Registration Error (rTRE) measures the Euclidean distance between corresponding landmarks after registration, normalized by the length of the image diagonal \( d_j \). It provides a dimensionless metric to evaluate the accuracy of landmark registration.

\begin{equation}
\text{rTRE}_{ij,l} = \frac{ \left\| \hat{\mathbf{x}}_{j,l} - \mathbf{x}_{j,l} \right\|_2 }{ d_j }
\label{eq:rTRE}
\end{equation}

where:
\[d_j = \sqrt{N_{x}^{j^2} + N_{y}^{j^2}}\] is the diagonal length of image \( j \), with \( N_{x}^{j} \) and \( N_{y}^{j} \) representing the dimensions of image \( j \) in the \( x \) and \( y \) directions, respectively. Here, \( \hat{\mathbf{x}}_{j,l} \) is the estimated position of landmark \( l \) in image \( j \) after registration, and \( \mathbf{x}_{j,l} \) is the ground truth position of the same landmark.

\noindent\textbf{Average Mean relative Target Registration Error (AMrTRE)} calculates the mean of the median rTRE values across all test image pairs. This metric provides an overall assessment of the registration accuracy by aggregating the central tendency of registration errors.

\begin{equation}
\text{AMrTRE} = \frac{1}{|T|} \sum_{(i,j) \in T} \mu^{i,j}
\label{eq:AMrTRE}
\end{equation}

where:
\[
\mu^{i,j} = \text{median}_{l \in L^{i,j}} \left( \text{rTRE}_{ij,l} \right)
\]
is the median rTRE for image pair \( (i,j) \). Here, \( T \) represents the set of all test image pairs, and \( L^{i,j} \) is the set of landmarks present in both images \( i \) and \( j \).

\noindent\textbf{Median of Median relative Target Registration Error (MMrTRE)} computes the median of the median rTRE values across all test image pairs. This metric offers a robust measure of registration accuracy by mitigating the influence of outliers.

\begin{equation}
\text{MMrTRE} = \text{median}_{(i,j) \in T} \left( \mu^{i,j} \right)
\label{eq:MMrTRE}
\end{equation}

\noindent\textbf{Average Maximum relative Target Registration Error (AMxrTRE)} calculates the mean of the maximum rTRE values for each test image pair. This metric highlights the average worst-case errors across the dataset, ensuring that extreme registration errors are accounted for.

\begin{equation}
\text{AMxrTRE} = \frac{1}{|T|} \sum_{(i,j) \in T} \left( \max_{l \in L^{i,j}} \text{rTRE}_{ij,l} \right)
\label{eq:AMxrTRE}
\end{equation}

\noindent\textbf{Average Registration Robustness (\( R_{\text{avg}} \))} measures the proportion of landmarks for which the registration error decreased compared to the initial error before registration. This metric evaluates the consistency of the registration method in improving landmark registration.

\begin{equation}
R_{\text{avg}} = \frac{1}{|T|} \sum_{(i,j) \in T} R^{i,j}
\label{eq:R_avg}
\end{equation}

where:
\[
R^{i,j} = \frac{ | K^{i,j} | }{ | L^{i,j} | }
\]
is the robustness for image pair \( (i,j) \), and
\[
K^{i,j} = \left\{ l \in L^{i,j} \mid \text{rTRE}_{ij,l} < \text{rIRE}_{ij,l} \right\}
\]
is the set of successfully registered landmarks for image pair \( (i,j) \). Here, \( \text{rIRE}_{ij,l} \) is the Relative Initial Registration Error before registration, defined as:
\[
\text{rIRE}_{ij,l} = \frac{ \left\| \mathbf{x}_{i,l} - \mathbf{x}_{j,l} \right\|_2 }{ d_j }
\]
where \( \mathbf{x}_{i,l} \) is the initial position of landmark \( l \) in image \( i \), assumed to be a reasonable approximation for \( \mathbf{x}_{j,l} \).

\noindent\textbf{Average Mean Distance (AMean\_D)} measures the average Euclidean distance between corresponding landmarks after registration, without normalization. This metric provides an absolute measure of registration error.

\begin{equation}
\text{AMean\_D} = \frac{1}{|T|} \sum_{(i,j) \in T} \left( \frac{1}{|L^{i,j}|} \sum_{l \in L^{i,j}} \left\| \hat{\mathbf{x}}_{j,l} - \mathbf{x}_{j,l} \right\|_2 \right)
\label{eq:AMean_distance}
\end{equation}

\subsection{Experiment Detail}

All experiments were conducted on the same workstation with a 48 GB Nvidia RTX A6000. The registration image sequence was processed using 3D Slicer to generate 3D visualizations. The image spacing was set to 1 $mm$ $\times$ 1 $mm$ $\times$ 8 $mm$ corresponding to a pixel size of 0.25 $\mu m$ under 40 $\times$ magnification.

\begin{figure*}[t]
\centering 
\includegraphics[width=1\linewidth]{{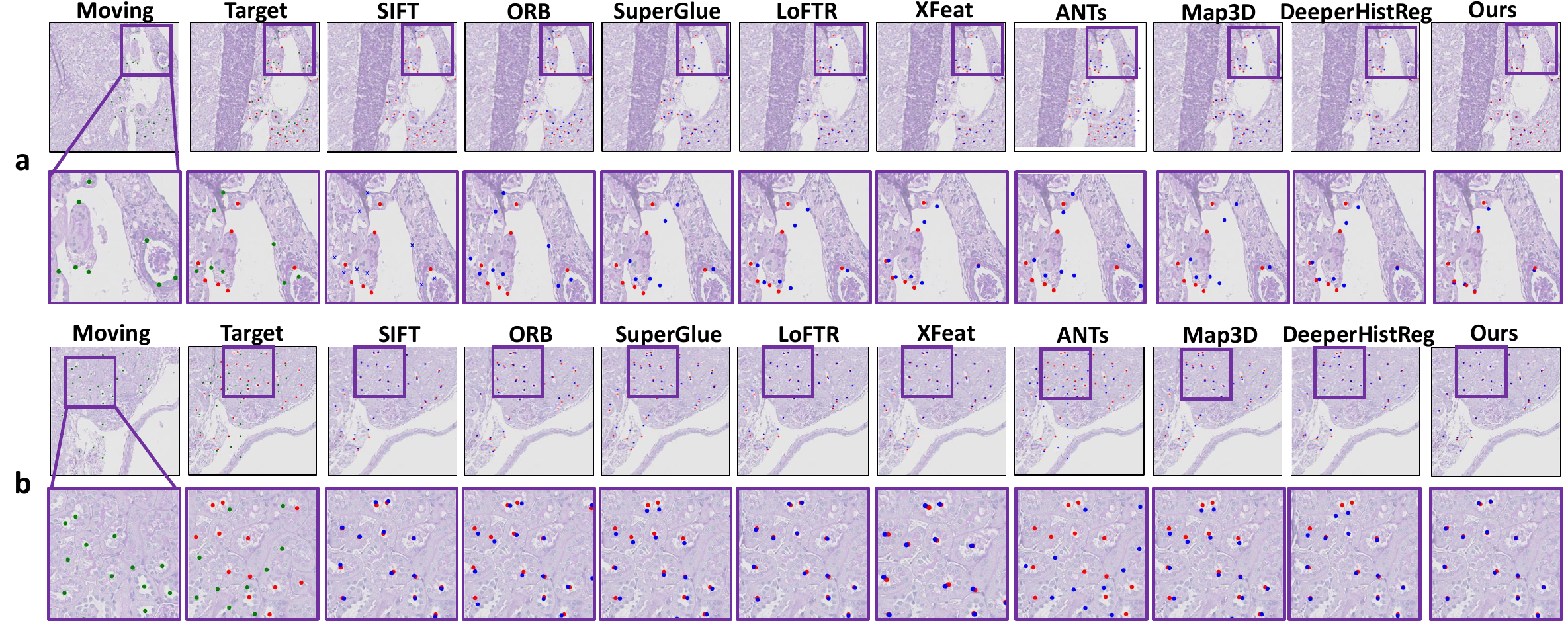}}
\caption{\textbf{Qualitative Result.} This figure showed  quantitative result between baseline method and our method. Part (a) shows the registration result on normal slice, part (b) shows the registration result on db/db diabetic slice. Green points represent landmarks in the moving image, red points indicate landmarks in the reference image, and blue points denote the transformed landmarks after registration. The proximity between blue and red points reflects the registration quality, with closer registration indicating better performance.} 
\label{result} 
\end{figure*}

\begin{table*}[ht]
\centering
\caption{Comparison of Registration Methods on Normal Mice Slice (Unit $\mu m$)}
\resizebox{\textwidth}{!}{%
\begin{tabular}{lcccccc}
\toprule
\textbf{Method} & \multicolumn{1}{c}{\textbf{AMrTRE}\,\(\downarrow\)} & \multicolumn{1}{c}{\textbf{MMrTRE}\,\(\downarrow\)} & \multicolumn{1}{c}{\textbf{AMean\_rTRE}\,\(\downarrow\)} & \multicolumn{1}{c}{\textbf{AMxrTRE}\,\(\downarrow\)} & \multicolumn{1}{c}{\textbf{R\_avg}\,\(\uparrow\)} & \multicolumn{1}{c}{\textbf{AMean\_D}\,\(\downarrow\)} \\
\midrule
SuperGlue\cite{DBLP:journals/corr/abs-1911-11763}     & 0.0263 & 0.0066 & 0.0275 & 0.0671 & 0.8738 & 19.9540 \\
LoFTR\cite{DBLP:journals/corr/abs-2104-00680}         & 0.0225 & 0.0048 & 0.0244 & 0.0527 & 0.8827 & 17.6954 \\
XFeat\cite{potje2024xfeatacceleratedfeatureslightweight}        & 0.0163 & 0.0049 & 0.0173 & 0.0383 & 0.9052 & 12.5473 \\
SIFT\cite{Lowe2004DistinctiveIF}           & 0.0124 & 0.0106 & 0.0149 & 0.0395 & 0.7379 & 10.8311 \\
ORB\cite{6126544}           & 0.0211 & 0.0115 & 0.0227 & 0.0534 & 0.7008 & 16.4771 \\
ANTs\cite{avants2009advanced}           & 0.0527 & 0.0366 & 0.0532 & 0.0807 & 0.3214 & 38.5389 \\
Map3D\cite{deng2021map3dregistrationbasedmultiobject}          & 0.0268 & 0.0078 & 0.0278 & 0.0679 & 0.8392 & 20.1966 \\
DeeperHistReg\cite{wodzinski2024deeperhistregrobustslideimages}  & 0.0026 & 0.0025 & 0.0043 & 0.0213 & 0.9726 & 3.1353 \\
\textbf{ZeroReg3D (Ours)}  & \textbf{0.0024} & \textbf{0.0024} & \textbf{0.0037} & \textbf{0.0177} & \textbf{0.9872} & \textbf{2.6952} \\
\bottomrule
\end{tabular}
}
\label{table1}
\end{table*}

\begin{table*}[ht]
\centering
\caption{Comparison of Registration Methods on db/db Diabetic Mice Slice (Unit $\mu m$)}
\resizebox{\textwidth}{!}{%
\begin{tabular}{lcccccc}
\toprule
\textbf{Method} & \multicolumn{1}{c}{\textbf{AMrTRE}\,\(\downarrow\)} & \multicolumn{1}{c}{\textbf{MMrTRE}\,\(\downarrow\)} & \multicolumn{1}{c}{\textbf{AMean\_rTRE}\,\(\downarrow\)} & \multicolumn{1}{c}{\textbf{AMxrTRE}\,\(\downarrow\)} & \multicolumn{1}{c}{\textbf{R\_avg}\,\(\uparrow\)} & \multicolumn{1}{c}{\textbf{AMean\_D}\,\(\downarrow\)} \\
\midrule
SuperGlue\cite{DBLP:journals/corr/abs-1911-11763}        & 0.0160 & 0.0051 & 0.0160 & 0.0375 & 0.8537 & 11.6208 \\
LoFTR\cite{DBLP:journals/corr/abs-2104-00680}           & 0.0147 & 0.0036 & 0.0170 & 0.0465 & 0.9122 & 12.3183 \\
XFeat\cite{potje2024xfeatacceleratedfeatureslightweight}            & 0.0160 & 0.0038 & 0.0156 & 0.0374 & 0.8964 & 11.3080 \\
SIFT\cite{Lowe2004DistinctiveIF}    & 0.0248 & 0.0086 & 0.0256 & 0.0548 & 0.5719 & 18.5785 \\
ORB\cite{6126544}              & 0.0262 & 0.0101 & 0.2708 & 0.0568 & 0.5965 & 19.6097 \\
ANTs\cite{avants2009advanced}             & 0.0438 & 0.0295 & 0.0441 & 0.0652 & 0.2882 & 31.9505 \\
Map3D\cite{deng2021map3dregistrationbasedmultiobject}            & 0.0249 & 0.0119 & 0.0247 & 0.0444 & 0.5265 & 17.8974 \\
DeeperHistReg\cite{wodzinski2024deeperhistregrobustslideimages}    & 0.0064 & \textbf{0.0025} & 0.0068 & \textbf{0.0218} & 0.9464 & 4.9953 \\
\textbf{ZeroReg3D (Ours)}    & \textbf{0.0030} & 0.0027 & \textbf{0.0059} & 0.0273 & \textbf{0.9754}  & \textbf{4.2919} \\
\bottomrule
\end{tabular}
}
\label{table2}
\end{table*}

%%%%%%%%%%%%%%%%% Human Fig and Table  %%%%%%%%%%%%%%%%%

\begin{figure*}[t]
\centering 
\includegraphics[width=1\linewidth]{{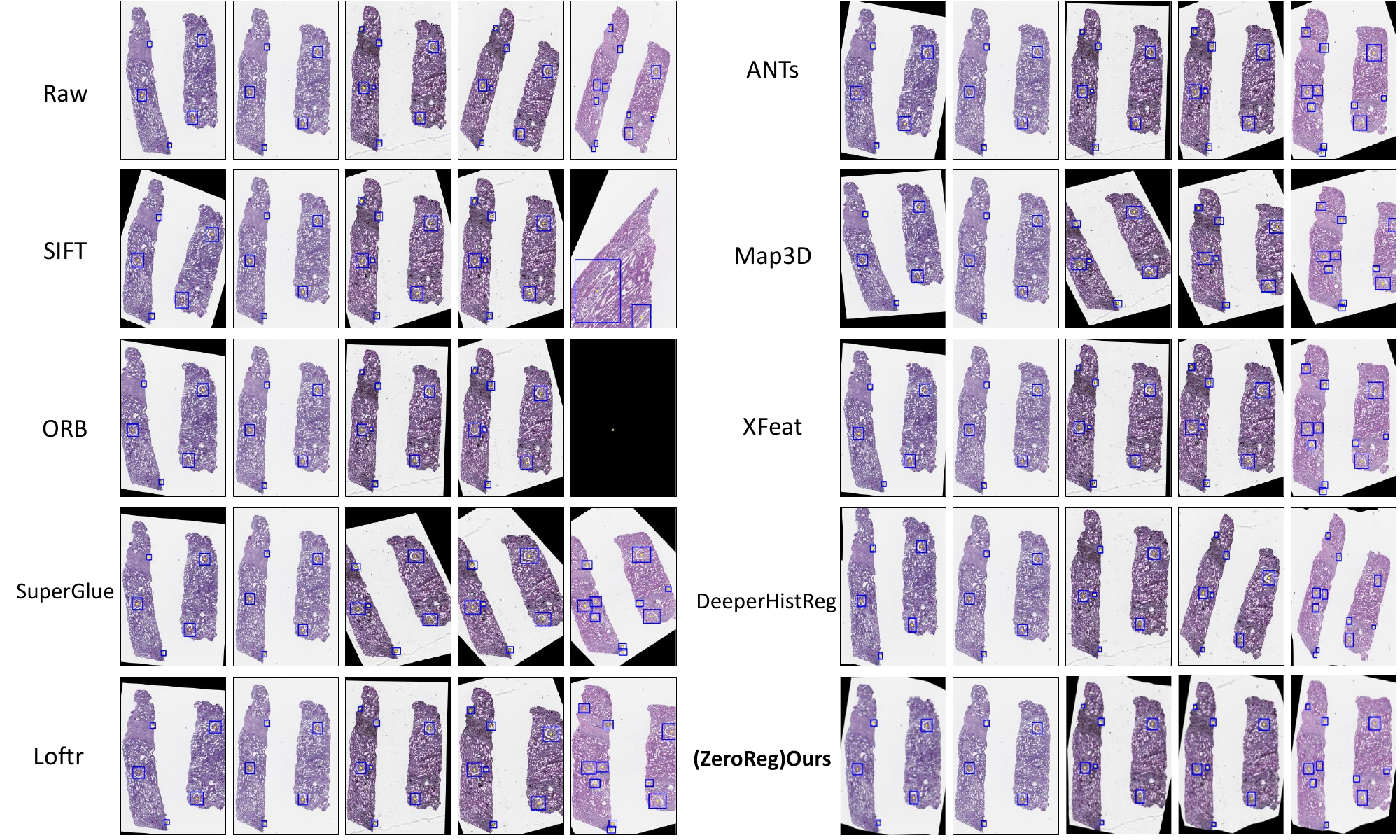}}
\caption{\textbf{Qualitative Result.} This figure showed qualitative result between baseline method and our method on human whole slide images datasets. Each yellow point indicates the center points of the glomeruli and the blue bounding box shows the size of the glomeruli.} 
\label{human_result} 
\end{figure*}

\begin{figure*}[t]
\centering 
\includegraphics[width=1\linewidth]{{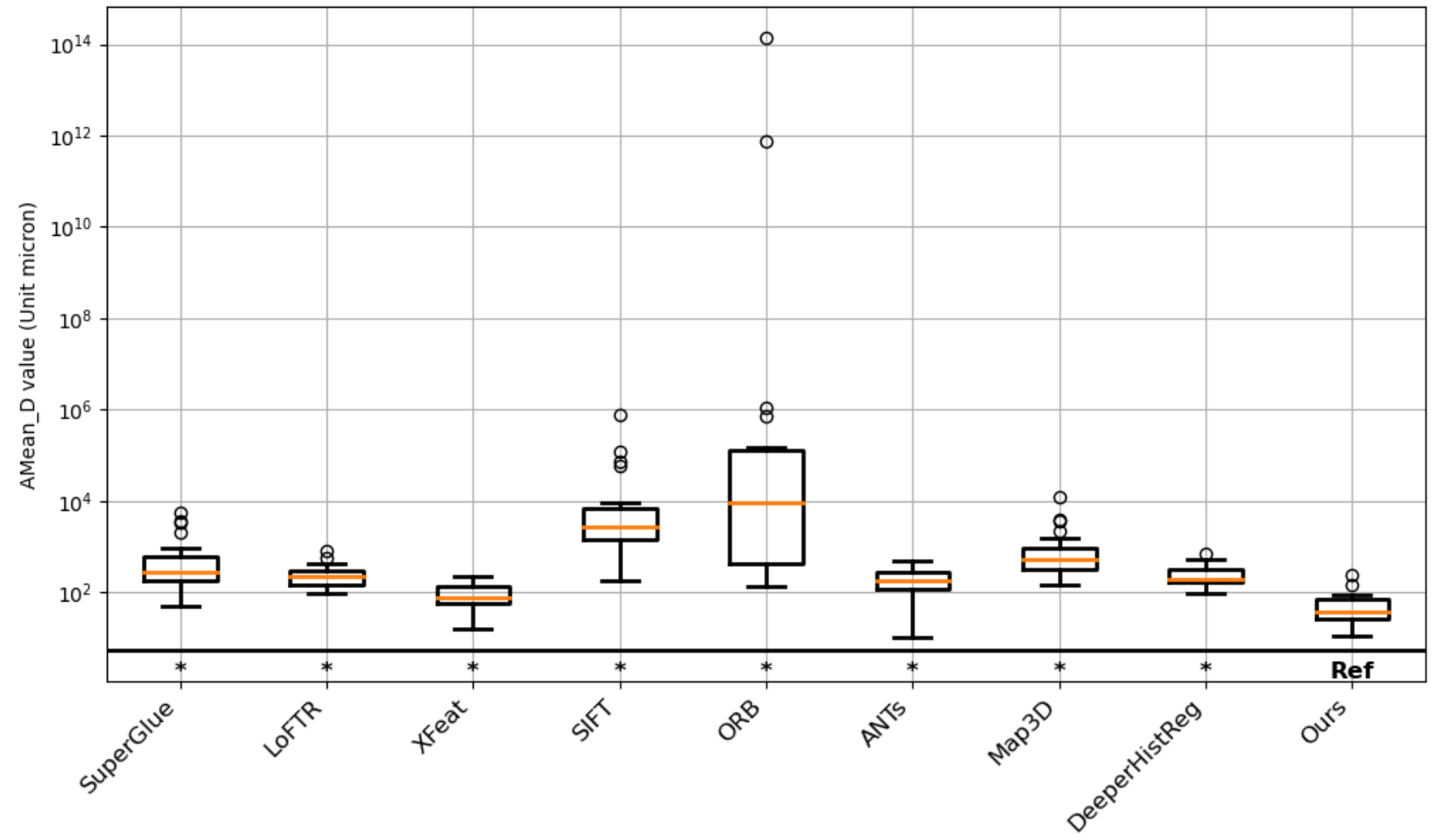}}
\caption{\textbf{Statistical Evaluation.} This figure shows statistical significance evaluation between baseline methods and our method on human whole slide image datasets. The Wilcoxon signed-rank test (\textit{p} $<$ 0.05) was used to assess statistical significance, with significant methods marked by an asterisk (*), indicating that our method consistently outperforms the other methods.} 
\label{box} 
\end{figure*}

\begin{table*}[ht]
\centering
\caption{Keypoints count before and after rotation preprocessing, along with timing statistics.}
\resizebox{\textwidth}{!}{%
\begin{tabular}{lccccc}
\hline
\textbf{Dataset} & \textbf{Slice Number} & \textbf{Keypoint Pairs (Before)} & \textbf{Keypoint Pairs (After)} & \textbf{Time (s)} & \textbf{Time per Slice (s)} \\
\hline

Mice & 66  & 12{,}965  & 16{,}882   & 5{,}030   & 76  \\
Human & 505 & 289{,}529 & 323{,}029 & 142{,}380 & 281 \\
\hline
\end{tabular}
}

\label{table3}
\end{table*}

%%%%%%%%%%%%%%%%%%%%%%%%%%%%%%%%%%%%%%%%%%
\section{Results}

Across the two mice datasets, our method consistently outperforms the baseline approaches in registration accuracy and robustness. As shown in ~\Cref{result}, the qualitative results clearly demonstrate that our approach achieves lower error metrics across a variety of challenging scenarios compared to existing methods. This improvement is evident across different evaluation metrics, highlighting the adaptability and precision of our registration framework.

Similarly, the evaluation on human datasets (see~\Cref{human_result}) reinforces the effectiveness of our method. The results indicate that our approach reliably produces lower AMean\_D values across multiple test cases.

Specifically, in the first mice dataset, our method achieves the lowest AMrTRE (\(0.0024\)) and AMean\_D (\(2.6952\)), indicating superior precision in aligning landmarks. Additionally, it records a high \( R\_{\text{avg}} \) (\(0.9872\)), demonstrating robust performance in consistently improving landmark registration as shown in Table~\ref{table1}. 

In the second mice dataset, our method maintains its leading performance with an AMrTRE of \(0.0030\) and an AMean\_distance of \(17.1674\). While some methods like xfeat also demonstrate competitive AMrTRE (\(0.0160\)) and AMean\_D (\(11.3080\)), our method still achieves the lowest values, underscoring its effectiveness. Notably, DeeperHistReg shows commendable performance with low AMrTRE (\(0.0064\)) and AMean\_distance (\(4.9953\)), but our method surpasses it with even lower error metrics as shown in Table~\ref{table2}.

The human datasets shown in Figure \ref{box} compare the AMean\_D values across nine methods over 20 human cases, with statistical significance assessed using the Wilcoxon signed-rank test. Our method exhibits the lowest median and least variability, indicating better and more stable performance than other methods. In contrast, SuperGlue, ORB, and SIFT show higher medians and larger spreads, suggesting greater variability and higher AMean\_D values. Our method is significantly better than baseline methods on this dataset.

Furthermore, several classical methods such as SIFT and ORB yield higher errors or exhibit missing values in multiple cases, underscoring their limitations in handling complex human datasets. In contrast, our method provides robust performance across all test cases and maintains competitive accuracy even in challenging scenarios where other advanced techniques falter.

We also evaluated the computational cost for the rotation preprocessing. For the mice dataset, which contains 66 slices, the number of keypoint pairs increased from 12,965 to 16,882 after applying the rotation step, with an average processing time of 76 seconds per slice. In the human dataset, consisting of 505 slices, the keypoint pairs increased from 289,529 to 323,029, with an average processing time of 281 seconds per slice.

% We aslo evaluated the performance of two distinct non-rigid image registration strategies on a globally rigidly aligned image sequence, using the central frame as the reference. Our method in the upper panel implemented pairwise non-rigid registration between consecutive frames followed by cumulative stacking, whereas the lower one involved directly registering each frame to the reference image. Given the inherent difficulty in establishing cross-pairwise landmarks, we employed 3D reconstructions of the registered sequences as the primary metric for assessing registration quality. Since it is difficult to find common landmarks throughout the entire dataset, we can not do the quantitative comparison between each method. The 3D visualization results unequivocally demonstrated that our method outperformed the other, producing higher-fidelity 3D reconstructions that indicate more precise and robust registration shown in ~\Cref{ablation}. 

% \begin{figure}[t]
% \centering 
% \includegraphics[width=1\linewidth]{{fig5.pdf}}
% \caption{\textbf{Ablation Study.} This figure shows two global-wise non-rigid registration methods. The upper panel shows the cascade registration results to the reference image and the lower panel shows the results that images directly register to the reference image.} 
% \label{ablation} 
% \end{figure}

\subsection{Discussion}

In this study, we introduced ZeroReg3D, a novel zero-shot registration pipeline designed to achieve precise and robust 3D reconstructions from serial histological sections. The superior performance observed across both mouse and human datasets can primarily be attributed to the careful integration of zero-shot deep learning-based keypoint extraction (XFeat) with robust affine and B-spline non-rigid registration methods. By combining deep learning techniques that inherently generalize well across diverse datasets with established mathematical models for deformation correction, our pipeline effectively mitigates common issues such as tissue deformation, sectioning artifacts, and variability in staining and illumination.

Specifically, the rotation preprocessing enhanced initial keypoint matching by ensuring optimal image orientation, thereby improving downstream affine and non-rigid transformations. The subsequent affine registration provided stable global registration, while the B-spline-based non-rigid registration captured subtle local tissue distortions, further refining the accuracy of registration. This holistic integration explains the consistent superiority of our method across various metrics, including lower AMrTRE, AMean\_D, and higher average robustness scores compared to existing state-of-the-art methods.

In terms of computational cost on the mouse dataset, ORB and SIFT required about 50s and 730s in total, respectively, while newer methods, LoFTR, a transformer-based matcher, and SuperGlue, a graph neural network-based matcher, took roughly 110s and 40s. XFeat, a lightweight feature extractor, further reduced total matching time to around 31s. Complete registration frameworks such as ANTs and DeeperHistReg then add their optimization overhead, requiring approximately 300s and 260s in total. Our complete pipeline uses XFeat matching in a single pass, completing in about 260 seconds, which is comparable to Map3D’s 400 seconds total. These results highlight that zero-shot matching with XFeat provides a favorable balance of accuracy and speed for mouse histology, making our approach especially practical for large-scale or resource-constrained studies.

Despite the strong performance, our method has limitations. One notable challenge is the computational cost associated with rotation preprocessing, which increases processing time. Additionally, handling substantial discontinuities between histological sections, particularly observed in human datasets where the inclusion of unstained slides resulted in significant structural gaps, posed challenges. These gaps complicated keypoint correspondence and subsequent non-rigid registration, occasionally leading to suboptimal reconstructions. Another limitation arises from potential inaccuracies introduced during cumulative stacking of pairwise registration, especially in datasets with many serial sections, potentially propagating registration errors along the sequence.

Furthermore, compared to patch-level registration, WSI level registration presents greater challenges. For instance, the typical diameter of mouse glomeruli is approximately 50 $\mu m$, with registration errors under 10 $\mu m$. In contrast, human glomeruli are larger—about 100 $\mu m$ in diameter—but exhibit higher registration errors of about 50 $\mu m$. One possible explanation is that mouse datasets generally consist of 20 to 40 serial sections, whereas human datasets often contain around 30 to 60 serial sections, increasing the likelihood of cumulative registration errors. Another contributing factor is the presence of tissue drift, fragmentation, or missing regions in the serial WSIs, which further complicates the registration process.

While our current study focuses on kidney tissue, it may be worthwhile to extend the proposed method to other tissue or organ types in the future studies. Future research will focus on addressing these limitations through several avenues. First, developing a rotation-invariant method could eliminate the need for rotation preprocessing, substantially reducing processing time. Additionally, building an end-to-end registration pipeline could decrease the interpolation steps from two to one, further improving computational efficiency and potentially enhancing registration quality.  Furthermore, we plan to explore global optimization methods that minimize cumulative registration errors over the entire sequence, potentially employing iterative refinement or joint registration techniques to enhance reconstruction fidelity.

\section{Conclusion}
In this study, we introduced ZeroReg3D, a novel zero-shot registration pipeline tailored for accurate 3D reconstruction from serial histological sections. By combining zero-shot deep learning-based keypoint matching with optimization-based affine and non-rigid registration techniques, ZeroReg3D effectively addresses critical challenges such as tissue deformation, sectioning artifacts, staining variability, and inconsistent illumination without requiring retraining or fine-tuning. Comprehensive evaluations demonstrated that our proposed pairwise 2D image registration method achieves approximately 10\% improvement in registration accuracy compared to state-of-the-art baseline methods, highlighting its superior precision and robustness. This framework provides a solid foundation for subsequent analysis transitioning from 2D serial sections to 3D histological tissue imaging.


\section*{Acknowledgment}
This research was supported by NIH R01DK135597 (Huo), DoD HT9425-23-1-0003 (HCY), and KPMP Glue Grant. This work was also supported by Vanderbilt Seed Success Grant, Vanderbilt Discovery Grant, and VISE Seed Grant. This project was supported by The Leona M. and Harry B. Helmsley Charitable Trust grant G-1903-03793 and G-2103-05128. This research was also supported by NIH grants R01EB033385, R01DK132338, REB017230, R01MH125931, and NSF 2040462. We extend gratitude to NVIDIA for their support by means of the NVIDIA hardware grant. This work was also supported by NSF NAIRR Pilot Award NAIRR240055.



%\bibliographystyle{IEEEtran}


% \end{thebibliography}

\bibliography{reference} % bibliography data in report.bib

\begin{thebibliography}{10}
\providecommand{\url}[1]{#1}
\csname url@samestyle\endcsname
\providecommand{\newblock}{\relax}
\providecommand{\bibinfo}[2]{#2}
\providecommand{\BIBentrySTDinterwordspacing}{\spaceskip=0pt\relax}
\providecommand{\BIBentryALTinterwordstretchfactor}{4}
\providecommand{\BIBentryALTinterwordspacing}{\spaceskip=\fontdimen2\font plus
\BIBentryALTinterwordstretchfactor\fontdimen3\font minus \fontdimen4\font\relax}
\providecommand{\BIBforeignlanguage}[2]{{%
\expandafter\ifx\csname l@#1\endcsname\relax
\typeout{** WARNING: IEEEtran.bst: No hyphenation pattern has been}%
\typeout{** loaded for the language `#1'. Using the pattern for}%
\typeout{** the default language instead.}%
\else
\language=\csname l@#1\endcsname
\fi
#2}}
\providecommand{\BIBdecl}{\relax}
\BIBdecl

\bibitem{xiong2024deep}
J.~Xiong, Y.~Liu, R.~Deng, R.~N. Tyree, H.~Correa, G.~Hiremath, Y.~Wang, and Y.~Huo, ``Deep learning-based open source toolkit for eosinophil detection in pediatric eosinophilic esophagitis,'' in \emph{Medical Imaging 2024: Digital and Computational Pathology}, vol. 12933.\hskip 1em plus 0.5em minus 0.4em\relax SPIE, 2024, pp. 231--237.

\bibitem{Zhu_2025_CVPR}
J.~Zhu, R.~Deng, T.~Yao, J.~Xiong, C.~Qu, J.~Guo, S.~Lu, M.~Yin, Y.~Wang, S.~Zhao, H.~Yang, and Y.~Huo, ``Asign: An anatomy-aware spatial imputation graphic network for 3d spatial transcriptomics,'' in \emph{Proceedings of the IEEE/CVF Conference on Computer Vision and Pattern Recognition (CVPR)}, June 2025, pp. 30\,829--30\,838.

\bibitem{deng2024hats}
R.~Deng, Q.~Liu, C.~Cui, T.~Yao, J.~Xiong, S.~Bao, H.~Li, M.~Yin, Y.~Wang, S.~Zhao, Y.~Tang, H.~Yang, and Y.~Huo, ``Hats: Hierarchical adaptive taxonomy segmentation for panoramic pathology image analysis,'' in \emph{International Conference on Medical Image Computing and Computer-Assisted Intervention}.\hskip 1em plus 0.5em minus 0.4em\relax Springer, 2024.

\bibitem{deng2024prpseg}
R.~Deng, Q.~Liu, C.~Cui, T.~Yao, J.~Yue, J.~Xiong, L.~Yu, Y.~Wu, M.~Yin, Y.~Wang \emph{et~al.}, ``Prpseg: Universal proposition learning for panoramic renal pathology segmentation,'' in \emph{Proceedings of the IEEE/CVF conference on computer vision and pattern recognition}, 2024, pp. 11\,736--11\,746.

\bibitem{Litjens}
G.~Litjens, C.~I. S{\'a}nchez, N.~Timofeeva, M.~Hermsen, I.~Nagtegaal, I.~Kovacs, C.~Hulsbergen-Van De~Kaa, P.~Bult, B.~Van~Ginneken, and J.~Van Der~Laak, ``Deep learning as a tool for increased accuracy and efficiency of histopathological diagnosis,'' \emph{Scientific reports}, vol.~6, no.~1, p. 26286, 2016.

\bibitem{Kather}
J.~N. Kather, A.~T. Pearson, N.~Halama, D.~J{\"a}ger, J.~Krause, S.~H. Loosen, A.~Marx, P.~Boor, F.~Tacke, U.~P. Neumann \emph{et~al.}, ``Deep learning can predict microsatellite instability directly from histology in gastrointestinal cancer,'' \emph{Nature medicine}, vol.~25, no.~7, pp. 1054--1056, 2019.

\bibitem{Brinker}
T.~J. Brinker, L.~Kiehl, M.~Schmitt, T.~B. Jutzi, E.~I. Krieghoff-Henning, D.~Krahl, H.~Kutzner, P.~Gholam, S.~Haferkamp, J.~Klode \emph{et~al.}, ``Deep learning approach to predict sentinel lymph node status directly from routine histology of primary melanoma tumours,'' \emph{European Journal of Cancer}, vol. 154, pp. 227--234, 2021.

\bibitem{Kiehl}
L.~Kiehl, S.~Kuntz, J.~H{\"o}hn, T.~Jutzi, E.~Krieghoff-Henning, J.~N. Kather, T.~Holland-Letz, A.~Kopp-Schneider, J.~Chang-Claude, A.~Brobeil \emph{et~al.}, ``Deep learning can predict lymph node status directly from histology in colorectal cancer,'' \emph{European Journal of Cancer}, vol. 157, pp. 464--473, 2021.

\bibitem{guo2024assessment}
J.~Guo, S.~Lu, C.~Cui, R.~Deng, T.~Yao, Z.~Tao, Y.~Lin, M.~Lionts, Q.~Liu, J.~Xiong \emph{et~al.}, ``Assessment of cell nuclei ai foundation models in kidney pathology,'' \emph{arXiv preprint arXiv:2408.06381}, 2024.

\bibitem{guo2024good}
------, ``How good are we? evaluating cell ai foundation models in kidney pathology with human-in-the-loop enrichment,'' \emph{arXiv preprint arXiv:2411.00078}, 2024.

\bibitem{zhu2025magnet}
J.~Zhu, R.~Deng, T.~Yao, J.~Xiong, C.~Qu, J.~Guo, S.~Lu, Y.~Tang, D.~Xu, M.~Yin \emph{et~al.}, ``Magnet: Multi-level attention graph network for predicting high-resolution spatial transcriptomics,'' \emph{arXiv preprint arXiv:2502.21011}, 2025.

\bibitem{xiong2024circle}
J.~Xiong, E.~H. Nguyen, Y.~Liu, R.~Deng, R.~N. Tyree, H.~Correa, G.~Hiremath, Y.~Wang, H.~Yang, A.~B. Fogo \emph{et~al.}, ``Circle representation for medical instance object segmentation,'' \emph{arXiv preprint arXiv:2403.11507}, 2024.

\bibitem{yue2025weighted}
J.~Yue, T.~Yao, R.~Deng, Q.~Liu, J.~Xiong, J.~Guo, H.~Yang, and Y.~Huo, ``Weighted circle fusion: ensembling circle representation from different object detection results,'' in \emph{Medical Imaging 2025: Image Perception, Observer Performance, and Technology Assessment}, vol. 13409.\hskip 1em plus 0.5em minus 0.4em\relax SPIE, 2025.

\bibitem{Chen}
Y.~Chen, Q.~Shen, S.~L. White, Y.~Gokmen-Polar, S.~Badve, and L.~J. Goodman, ``Three-dimensional imaging and quantitative analysis in clarity processed breast cancer tissues,'' \emph{Scientific reports}, vol.~9, no.~1, p. 5624, 2019.

\bibitem{Merz}
S.~F. Merz, P.~Jansen, R.~Ulankiewicz, L.~Bornemann, T.~Schimming, K.~Griewank, Z.~Cibir, A.~Kraus, I.~Stoffels, T.~Aspelmeier \emph{et~al.}, ``High-resolution three-dimensional imaging for precise staging in melanoma,'' \emph{European Journal of Cancer}, vol. 159, pp. 182--193, 2021.

\bibitem{Geng}
J.~Geng, X.~Zhang, S.~Prabhu, S.~H. Shahoei, E.~R. Nelson, K.~S. Swanson, M.~A. Anastasio, and A.~M. Smith, ``3d microscopy and deep learning reveal the heterogeneity of crown-like structure microenvironments in intact adipose tissue,'' \emph{Science advances}, vol.~7, no.~8, p. eabe2480, 2021.

\bibitem{Pichat}
J.~Pichat, J.~E. Iglesias, T.~Yousry, S.~Ourselin, and M.~Modat, ``A survey of methods for 3d histology reconstruction,'' \emph{Medical image analysis}, vol.~46, pp. 73--105, 2018.

\bibitem{saalfeld2012elastic}
S.~Saalfeld, R.~Fetter, A.~Cardona, and P.~Tomancak, ``Elastic volume reconstruction from series of ultra-thin microscopy sections,'' \emph{Nature methods}, vol.~9, no.~7, pp. 717--720, 2012.

\bibitem{auto3dreg2015}
A.~Author and B.~Coauthor, ``Fully automatic and robust 3d registration of serial-section microscopy images,'' \emph{Scientific Reports}, vol.~5, p. 15051, 2015.

\bibitem{lotz2021comparison}
J.~Lotz, N.~Weiss, J.~van~der Laak, and S.~Heldmann, ``Comparison of consecutive and re-stained sections for image registration in histopathology,'' \emph{arXiv preprint arXiv:2106.13150}, 2021.

\bibitem{potje2024xfeatacceleratedfeatureslightweight}
\BIBentryALTinterwordspacing
G.~Potje, F.~Cadar, A.~Araujo, R.~Martins, and E.~R. Nascimento, ``Xfeat: Accelerated features for lightweight image matching,'' 2024. [Online]. Available: \url{https://arxiv.org/abs/2404.19174}
\BIBentrySTDinterwordspacing

\bibitem{DBLP:journals/corr/abs-1911-11763}
\BIBentryALTinterwordspacing
P.~Sarlin, D.~DeTone, T.~Malisiewicz, and A.~Rabinovich, ``Superglue: Learning feature matching with graph neural networks,'' \emph{CoRR}, vol. abs/1911.11763, 2019. [Online]. Available: \url{http://arxiv.org/abs/1911.11763}
\BIBentrySTDinterwordspacing

\bibitem{jiang2024omnigluegeneralizablefeaturematching}
\BIBentryALTinterwordspacing
H.~Jiang, A.~Karpur, B.~Cao, Q.~Huang, and A.~Araujo, ``Omniglue: Generalizable feature matching with foundation model guidance,'' 2024. [Online]. Available: \url{https://arxiv.org/abs/2405.12979}
\BIBentrySTDinterwordspacing

\bibitem{DBLP:journals/corr/abs-2104-00680}
\BIBentryALTinterwordspacing
J.~Sun, Z.~Shen, Y.~Wang, H.~Bao, and X.~Zhou, ``Loftr: Detector-free local feature matching with transformers,'' \emph{CoRR}, vol. abs/2104.00680, 2021. [Online]. Available: \url{https://arxiv.org/abs/2104.00680}
\BIBentrySTDinterwordspacing

\bibitem{DeTone2016DeepIH}
D.~DeTone, L.~Malisiewicz, and A.~Rabinovich, ``Deep image homography estimation,'' in \emph{Proceedings of the IEEE Conference on Computer Vision and Pattern Recognition (CVPR)}, 2016.

\bibitem{466930}
P.~Viola and W.~Wells, ``Alignment by maximization of mutual information,'' in \emph{Proceedings of IEEE International Conference on Computer Vision}, 1995, pp. 16--23.

\bibitem{1660446}
F.~Zhao, Q.~Huang, and W.~Gao, ``Image matching by normalized cross-correlation,'' in \emph{2006 IEEE International Conference on Acoustics Speech and Signal Processing Proceedings}, vol.~2, 2006, pp. II--II.

\bibitem{1284395}
Z.~Wang, A.~Bovik, H.~Sheikh, and E.~Simoncelli, ``Image quality assessment: from error visibility to structural similarity,'' \emph{IEEE Transactions on Image Processing}, vol.~13, no.~4, pp. 600--612, 2004.

\bibitem{Lowe2004DistinctiveIF}
\BIBentryALTinterwordspacing
D.~G. Lowe, ``Distinctive image features from scale-invariant keypoints,'' \emph{International Journal of Computer Vision}, vol.~60, pp. 91--110, 2004. [Online]. Available: \url{https://api.semanticscholar.org/CorpusID:174065}
\BIBentrySTDinterwordspacing

\bibitem{6126544}
E.~Rublee, V.~Rabaud, K.~Konolige, and G.~Bradski, ``Orb: An efficient alternative to sift or surf,'' in \emph{2011 International Conference on Computer Vision}, 2011, pp. 2564--2571.

\bibitem{Rueckert1999NonrigidRU}
\BIBentryALTinterwordspacing
D.~Rueckert, L.~I. Sonoda, C.~Hayes, D.~L.~G. Hill, M.~O. Leach, and D.~J. Hawkes, ``Nonrigid registration using free-form deformations: application to breast mr images,'' \emph{IEEE Transactions on Medical Imaging}, vol.~18, pp. 712--721, 1999. [Online]. Available: \url{https://api.semanticscholar.org/CorpusID:330039}
\BIBentrySTDinterwordspacing

\bibitem{thirion1998image}
J.-P. Thirion, ``Image matching as a diffusion process: an analogy with maxwell's demons,'' \emph{Medical Image Analysis}, vol.~2, no.~3, pp. 243--260, 1998.

\bibitem{24792}
F.~Bookstein, ``Principal warps: thin-plate splines and the decomposition of deformations,'' \emph{IEEE Transactions on Pattern Analysis and Machine Intelligence}, vol.~11, no.~6, pp. 567--585, 1989.

\bibitem{avants2009advanced}
B.~B. Avants, N.~Tustison, G.~Song \emph{et~al.}, ``Advanced normalization tools (ants),'' \emph{Insight j}, vol.~2, no. 365, pp. 1--35, 2009.

\bibitem{deng2021map3dregistrationbasedmultiobject}
\BIBentryALTinterwordspacing
R.~Deng, H.~Yang, A.~Jha, Y.~Lu, P.~Chu, A.~B. Fogo, and Y.~Huo, ``Map3d: Registration based multi-object tracking on 3d serial whole slide images,'' 2021. [Online]. Available: \url{https://arxiv.org/abs/2006.06038}
\BIBentrySTDinterwordspacing

\bibitem{wodzinski2024deeperhistregrobustslideimages}
\BIBentryALTinterwordspacing
M.~Wodzinski, N.~Marini, M.~Atzori, and H.~Müller, ``Deeperhistreg: Robust whole slide images registration framework,'' 2024. [Online]. Available: \url{https://arxiv.org/abs/2404.14434}
\BIBentrySTDinterwordspacing

\bibitem{DBLP:journals/corr/abs-1809-05231}
\BIBentryALTinterwordspacing
G.~Balakrishnan, A.~Zhao, M.~R. Sabuncu, J.~V. Guttag, and A.~V. Dalca, ``Voxelmorph: {A} learning framework for deformable medical image registration,'' \emph{CoRR}, vol. abs/1809.05231, 2018. [Online]. Available: \url{http://arxiv.org/abs/1809.05231}
\BIBentrySTDinterwordspacing

\bibitem{9058666}
J.~Borovec, J.~Kybic, I.~Arganda-Carreras, D.~V. Sorokin, G.~Bueno, A.~V. Khvostikov, S.~Bakas, E.~I.-C. Chang, S.~Heldmann, K.~Kartasalo, L.~Latonen, J.~Lotz, M.~Noga, S.~Pati, K.~Punithakumar, P.~Ruusuvuori, A.~Skalski, N.~Tahmasebi, M.~Valkonen, L.~Venet, Y.~Wang, N.~Weiss, M.~Wodzinski, Y.~Xiang, Y.~Xu, Y.~Yan, P.~Yushkevich, S.~Zhao, and A.~Muñoz-Barrutia, ``Anhir: Automatic non-rigid histological image registration challenge,'' \emph{IEEE Transactions on Medical Imaging}, vol.~39, no.~10, pp. 3042--3052, 2020.

\end{thebibliography}
\bibliographystyle{IEEEtran} % 

\end{document}